\documentclass[letterpaper, 10 pt, conference]{ieeeconf}  

\IEEEoverridecommandlockouts                              

\usepackage{graphicx}
\usepackage{cite}
\usepackage{picinpar}
\usepackage{amsmath}
\usepackage{url}
\usepackage{flushend}
\usepackage{colortbl}
\usepackage{soul}
\usepackage{multirow}
\usepackage{pifont}
\usepackage{color}
\usepackage{alltt}
\usepackage[hidelinks]{hyperref}
\usepackage{enumerate}
\usepackage{siunitx}
\usepackage{epstopdf}
\usepackage{pbox}

\usepackage{amssymb}
\usepackage{stfloats}
\usepackage{float}

\usepackage{xcolor}
\usepackage[ruled,linesnumbered]{algorithm2e}
\usepackage{tabularx}
\usepackage{multirow}
\newcommand{\tabincell}[2]{\begin{tabular}{@{}#1@{}}#2\end{tabular}}

\DeclareSymbolFont{largesymbol}{OMX}{yhex}{m}{n}
\DeclareMathAccent{\Widehat}{\mathord}{largesymbol}{"62}

\title{\LARGE \bf
Autonomous Magnetic Navigation Framework for Active Wireless Capsule Endoscopy Inspired by Conventional Colonoscopy Procedures
}

\author{Yangxin~Xu$^{*}$,~\IEEEmembership{Student~Member,~IEEE,}
        Keyu~Li$^{*}$,~\IEEEmembership{Student~Member,~IEEE,}
        Ziqi~Zhao, \\
        and~Max~Q.-H.~Meng$^{\sharp}$,~\IEEEmembership{Fellow,~IEEE}
\thanks{This work is partially supported by National Key R \& D program of China with Grant No. 2019YFB1312400 and Hong Kong RGC CRF grant C4063-18G awarded to Max Q.-H. Meng.}
\thanks{Y. Xu and K. Li are with the Department of Electronic Engineering, the Chinese University of Hong Kong, Hong Kong SAR, China (e-mail: yxxu@link.cuhk.edu.hk; kyli@link.cuhk.edu.hk).}
\thanks{Z. Zhao is with the Department of Electronic and Electrical Engineering, the Southern University of Science and Technology, Shenzhen, China (e-mail: zzq2694@163.com).}
\thanks{Max Q.-H. Meng is with the Department of Electronic and Electrical Engineering of the Southern University of Science and Technology in Shenzhen, China, on leave from the Department of Electronic Engineering, the Chinese University of Hong Kong, Hong Kong SAR, China, and also with the Shenzhen Research Institute of the Chinese University of Hong Kong, Shenzhen, China (e-mail: max.meng@ieee.org).}
\thanks{$^{*}$ The authors contribute equally to this paper.}
\thanks{$^{\sharp}$ Corresponding author.}
}

\begin{document}

\maketitle
\thispagestyle{empty}
\pagestyle{empty}

%
%

\begin{abstract}
In recent years, simultaneous magnetic actuation and localization (SMAL) for active  wireless capsule endoscopy (WCE) has been intensively studied to improve the efficiency and accuracy of the examination.
In this paper, we propose an autonomous magnetic navigation framework for active WCE that mimics the ``insertion" and ``withdrawal" procedures performed by an expert physician in conventional colonoscopy, thereby enabling efficient and accurate navigation of a robotic capsule endoscope in the intestine with minimal user effort. First, the capsule is automatically propelled through the unknown intestinal environment and generate a viable path to represent the environment. Then, the capsule is autonomously navigated towards any point selected on the intestinal trajectory to allow accurate and repeated inspections of suspicious lesions. Moreover, we implement the navigation framework on a robotic system incorporated with advanced SMAL algorithms, and validate it in the navigation in various tubular environments using phantoms and an ex-vivo pig colon. Our results demonstrate that the proposed autonomous navigation framework can effectively navigate the capsule in unknown, complex tubular environments with a satisfactory accuracy, repeatability and efficiency compared with manual operation.
\end{abstract}

\section{INTRODUCTION}

\begin{figure}[tb]
\setlength{\abovecaptionskip}{-0.4cm}
\centering
\includegraphics[scale=1.0,angle=0,width=0.49\textwidth]{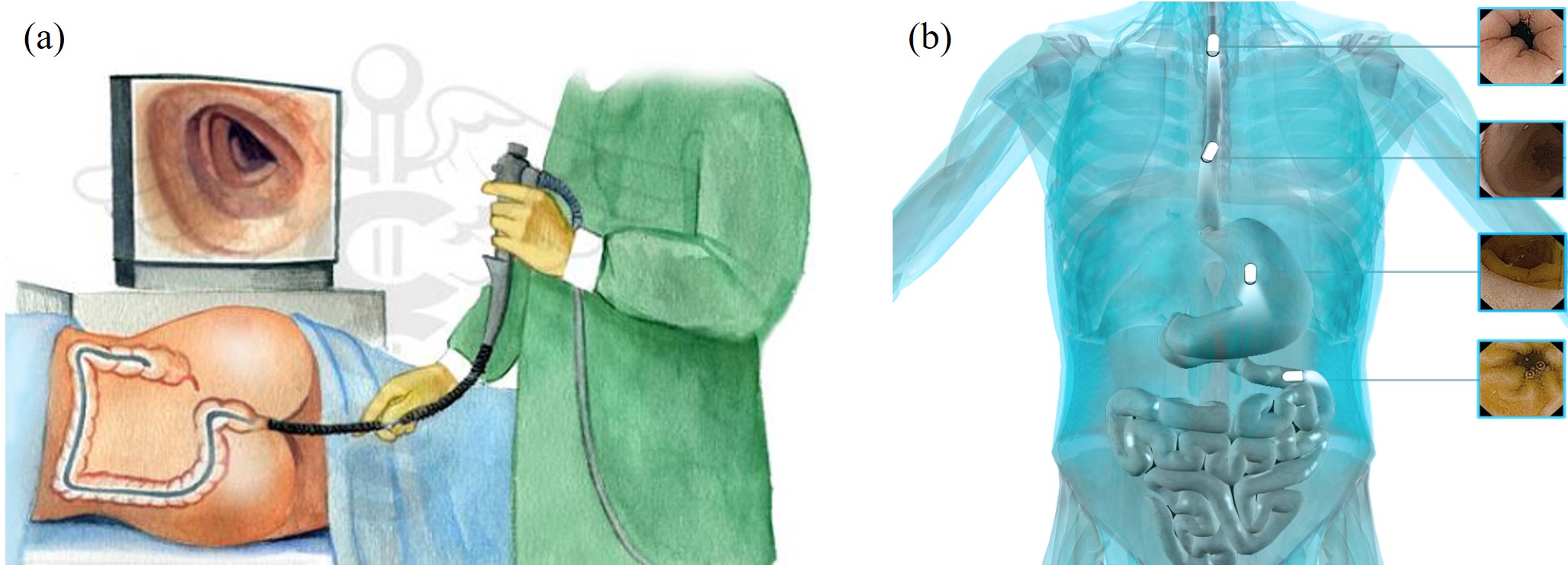}
\caption[Conventional colonoscopy and WCE]{(a) shows the procedure of conventional colonoscopy. (b) shows the wireless capsule endoscopy (WCE).}
\label{Fig_colonoscopy_WCE}
\end{figure}

Colorectal cancer is the development of cancer in the final segment of the gastrointestinal (GI) system, including the colon, rectum and anus \cite{siegel2020colorectal}\cite{sung2021global}. According to the Global Cancer Statistics 2020 \cite{sung2021global}, colorectal cancer has become the second leading cause of cancer death with an estimated 0.94 million deaths, accounting for $9.4\%$ of the total cancer deaths. Colonoscopy is the current ``gold standard" to control and prevent colorectal diseases by early diagnosis and early treatment \cite{Joseph2016Colorectal}. However, the commonly performed colonoscopy requires a trained clinician to manually insert the endoscope, as shown in Fig. \ref{Fig_colonoscopy_WCE}(a). This procedure usually causes patient pain and discomfort, and introduces safety risks such as tissue perforation \cite{martin2020enabling}. Moreover, since the procedure requires a long learning curve of the operator, the lack of skilled endoscopists and medical resources in rural areas has resulted in the urban-rural disparity in colorectal cancer incidence \cite{wen2018urban}.

To address these issues, wireless capsule endoscopy (WCE) was introduced in 2000 as a painless and non-invasive tool to inspect the entire GI tract \cite{iddan2000wireless}. As shown in Fig. \ref{Fig_colonoscopy_WCE}(b), The patient is required to undergo bowel cleansing and swallow a capsule containing tiny encapsulated cameras to captures images of the colon and rectum. However, the currently used WCE is passively pushed though the intestine by peristalsis, which results in a very long procedural time (the whole inspection of the GI tract takes about $8 \sim 24$ hours \cite{meng2004wireless}) and the flexibility of examination is limited as the capsule cannot be intuitively positioned \cite{meng2004wireless}.

Active WCE is a concept of endowing WCE with active locomotion and precise localization to achieve autonomous navigation in the GI tract, which holds great promise to realize painless, efficient and accurate diagnosis and therapy with minimal manual operations \cite{ciuti2011capsule}.
In recent years, simultaneous magnetic actuation and localization (SMAL) technologies have been intensively studied for active WCE, which utilize the magnetic fields to actuate and locate the capsule at the same time \cite{bianchi2019localization}\cite{shamsudhin2017magnetically}. The actuating magnetic field can be generated by electromagnetic coils or external permanent magnets \cite{abbott2020magnetic}, resulting in different system design and control strategies.
In view of a clinical translation of the SMAL technologies for active WCE, two fundamental questions need to be answered: 
\begin{enumerate} 
\item \textit{How to automatically propel a capsule through an unknown intestinal environment?} 
\item \textit{How to accurately control the capsule to reach a given point?} 
\end{enumerate}

The first task focuses on the fast exploration of an unknown tubular environment by automatically propelling the robotic capsule, which we refer to as the ``automatic propulsion" (AP) task; while the second task is similar to the trajectory following problem, where the environment is assumed known and the capsule is required to accurately follow a pre-defined trajectory, which we refer to as the ``trajectory following" (TF) task. 

Previous studies have separately dealt with the above two tasks. Several SMAL systems have been developed to address the AP task based solely on the magnetic feedback\cite{popek2017first,xu2020novelsystem,xu2019towards,xu2020improved} or on a fusion of magnetic and visual feedback \cite{martin2020enabling}. While these methods can determine the real-time movement of the capsule based on the estimated direction of the intestine, the actual moving trajectory of the capsule cannot be precisely controlled to allow repeated inspections at given points (e.g., suspicious lesions) for high-quality diagnosis.
Other groups have exclusively focused on the TF task of a capsule to accurately follow manually defined trajectories \cite{norton2019intelligent,pittiglio2019magnetic,taddese2018enhanced,barducci2019adaptive,scaglioni2019explicit}. However, the human intestinal environment is complex and unconstructed due to patient variability and unknown obstacles, so the simple, pre-defined trajectories may not be viable during clinical applications and would limit the flexibility of the examination.

In this work, we go beyond just investigating the AP or TF task for active WCE, and answer the two fundamental questions by providing a general framework for autonomous navigation of a capsule in unknown tubular environments. Our method is inspired by conventional colonoscopy procedures and combines efficient exploration of an unknown tubular environment and accurate tracking of given trajectories through a workflow that mimics the skills of an expert colonoscopist. Moreover, we develop a SMAL system based on an external sensor array and a reciprocally rotating magnetic actuator to implement the autonomous navigation framework, in order to achieve safe, efficient and accurate navigation of a capsule in the intestine. Our proposed framework is validated in real-world experiments on both phantoms and ex-vivo pig colons. The contributions are summarized as follows:

\begin{itemize}
\item An autonomous navigation framework for active WCE is first proposed in this paper, which mimics an expert colonoscopist performing the ``insertion" and ``withdrawal" procedures in routine colonoscopy to achieve efficient exploration of an unknown tubular environment and accurate inspection of suspicious lesions with minimal user effort.
\item The proposed framework is implemented on a real robotic system, which is incorporated with adaptive magnetic localization and reciprocally rotating magnetic actuation-based AP and TF algorithms, and is validated in extensive navigation experiments in various tubular environments.
\item A thorough analysis of the experiment results is provided to compare different methods used in our framework in different tasks. The results demonstrate that our autonomous magnetic navigation method can achieve comparable accuracy and improved repeatability and efficiency compared to manual control.
\end{itemize}

\section{METHODS}
In this section, we first introduce the concept of autonomous navigation for active WCE that mimics the skills of an expert clinician in routine colonoscopy, and present a general workflow to realize efficient exploration and accurate inspection of the intestine with minimal user effort. Then, we introduce the magnetic actuation methods and describe the system and algorithms we use to implement the framework to navigate a capsule in unknown tubular environments.

\subsection{Autonomous Navigation Framework for Active WCE}

\begin{figure}[t]
\setlength{\abovecaptionskip}{-0.4cm}
\centering
\includegraphics[scale=1.0,angle=0,width=0.49\textwidth]{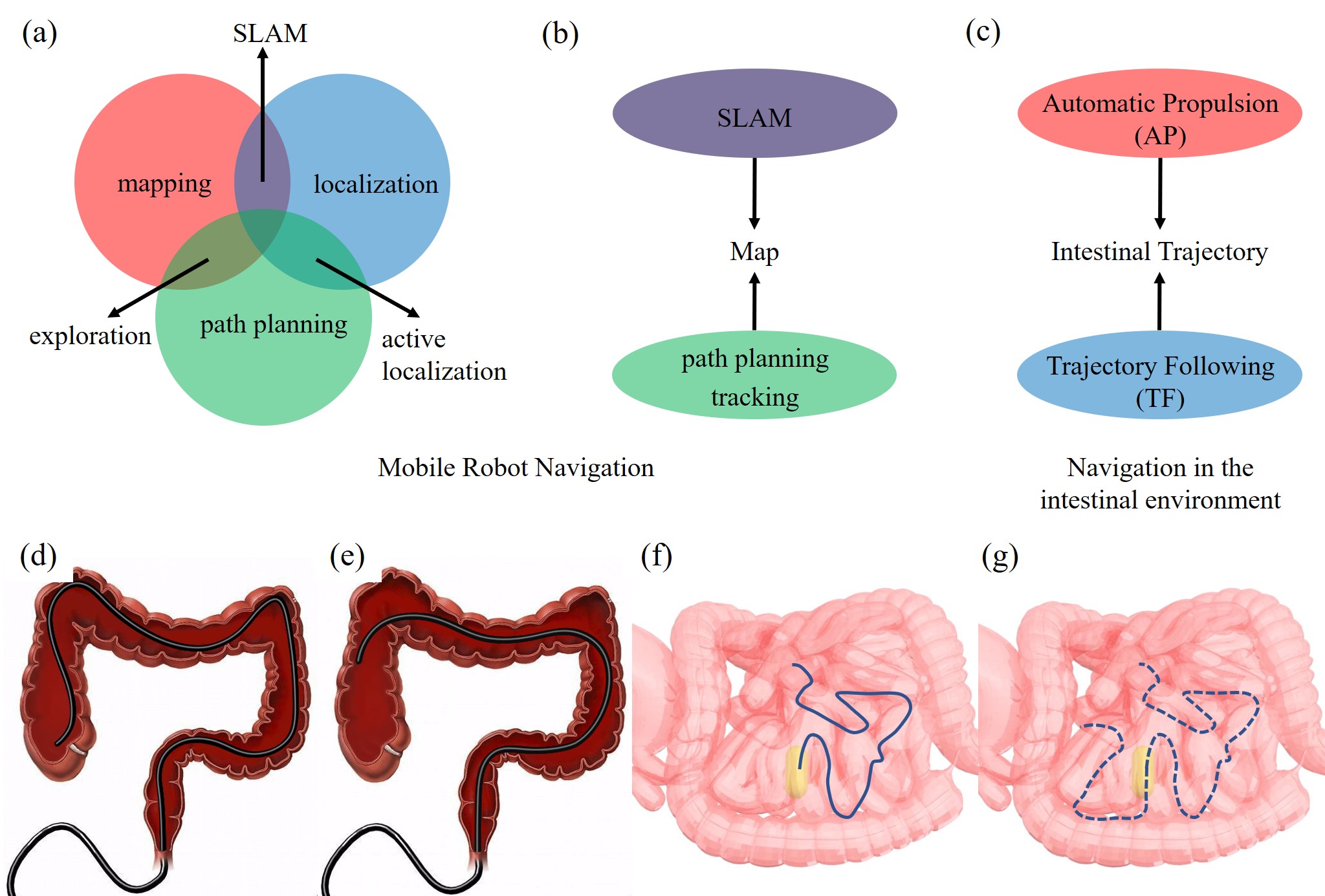}
\caption[The proposed autonomous navigation framework]{(a) shows the classic concepts in mobile robot navigation. (b) is a simplified version of (a). (c) shows our proposed concept for WCE navigation in the intestinal environment. (d-e) show the ``insertion" and ``withdrawal" steps in conventional colonoscopy, respectively. (f) During the AP step, the capsule (yellow) explores the unknown intestinal environment and a viable trajectory is generated (indigo line). (g) During the TF step, the capsule is navigated towards any selected point on the trajectory (dashed indigo line).}
\label{Fig_overview_mimicconcepts}
\end{figure}

The conventional colonoscopy requires an experienced physician to perform the ``insertion" and ``withdrawal" procedures to manipulate the endoscope during the examination\cite{lee2014colonoscopy}. As shown in Fig. \ref{Fig_overview_mimicconcepts}(d), the physician first pushes forward the endoscope from anus to cecum during the ``insertion" procedure to find a viable path in the unknown intestine and perform preliminary detection of abnormalities, which usually takes a long time due to the unknown friction and shape of the intestinal environment. Subsequently, the ``withdrawal" procedure is performed as shown in Fig. \ref{Fig_overview_mimicconcepts}(e), during which the endoscope can be smoothly pulled back through the intestine and the entire intestine can be inspected carefully for high-quality diagnosis. This motivates us to design a general workflow for the autonomous robotic navigation of active WCE that mimics the insertion-withdrawal procedures in conventional colonoscopy, to first perform efficient exploration in the unknown intestinal environment and then conduct accurate navigation towards suspicious lesions. In other words, the AP and TF tasks are executed successively in our navigation framework to realize coarse-to-fine navigation in an unknown tubular environment. 

To better illustrate the proposed method, we compare our autonomous navigation framework for active WCE with the classic concepts in mobile robot navigation, as shown in Fig. \ref{Fig_overview_mimicconcepts}(a-b), including the mapping, localization and path planning modules \cite{stachniss2009robotic}. After the map of an unknown environment is generated, the mobile robot can navigate in the map by executing a planned path under motion control. In comparison, in the navigation framework for WCE (see Fig. \ref{Fig_overview_mimicconcepts}(c)), during the AP step, the robotic capsule is actuated to explore the unknown intestinal environment, and a viable path is generated to represent the environment, as illustrated in Fig. \ref{Fig_overview_mimicconcepts}(f). In the TF step, the robotic capsule can accurately navigate towards any selected point on the trajectory with control strategies, as shown in Fig. \ref{Fig_overview_mimicconcepts}(g).

\begin{figure}[t]
\centering
\includegraphics[scale=1.0,angle=0,width=0.42\textwidth]{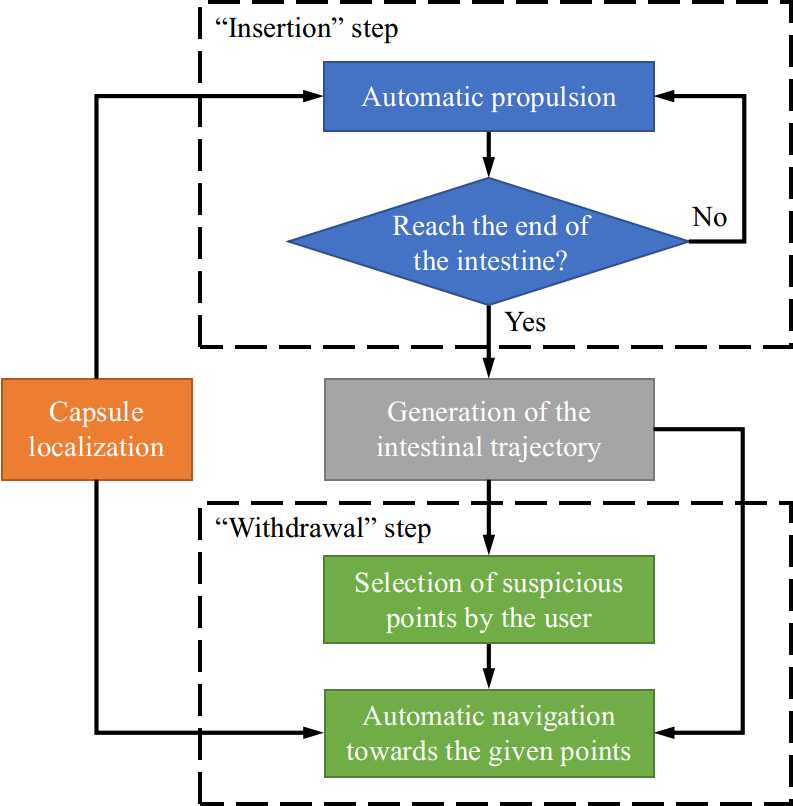}
\caption[Workflow of our proposed autonomous navigation framework]{Workflow of our proposed autonomous navigation framework for active WCE in a clinical setting. The capsule is actuated and localized with a robotic system to automate the ``insertion" and "withdrawal" procedures in routine colonoscopy. }
\label{Fig_overview_workflow}
\vspace{-0.4cm}
\end{figure}

In view of a clinical integration, the AP and TF steps can be executed by a robotic system with minimal user effort through the workflow summarized in Fig. \ref{Fig_overview_workflow}. Specifically, the capsule is first advanced through the entire intestine by the AP algorithm under the supervision of a physician. After the capsule reaches the end of the intestine (determined by the user), a smooth trajectory is generated to represent the intestine. This step mimics the ``insertion" step in routine colonoscopy. Then, the user can select a set of suspicious points on the trajectory, and the capsule will be automatically controlled to reach the selected points using the TF algorithm. This step is associated with the ``withdrawal" step in routine colonoscopy to facilitate accurate and repeated inspection of suspected lesions. Combining the advantages of the AP and TF techniques, this framework can realize both flexible and accurate navigation in the unknown intestinal environment, thereby enabling effective robotic capsule endoscopy with minimal manual operations.

\subsection{Magnetic Actuation Methods}
In this work, we consider a permanent magnet-based SMAL system to realize the autonomous navigation of the capsule.
The magnetic actuation methods applied in existing permanent magnet-based SMAL systems for active WCE can be roughly classified into three categories: i) \textit{Dragging magnetic actuation} (DMA), which directly uses the magnetic force generated by the magnetic actuator to drag or steer the capsule \cite{martin2020enabling}\cite{taddese2018enhanced}\cite{barducci2019adaptive}\cite{mahoney2016five}; ii) \textit{Continuously rotating magnetic actuation} (CRMA), which uses a rotating magnetic field generated by a continuously rotating magnetic actuator for helical propulsion of a capsule with external thread in a tubular environment \cite{xu2020novelsystem,xu2019towards,xu2020improved,mahoney2014generating,popek2016six,xu2020novel}, and iii) \textit{Reciprocally rotating magnetic actuation} (RRMA), which uses a reciprocally rotating magnetic actuator to rotate a non-threaded capsule back and forth during propulsion in a tubular environment \cite{xu2021reciprocally}\cite{xu2021reciprocallyTF}.
Since the RRMA method \cite{xu2021reciprocally} was introduced to reduce the risk of causing intestinal malrotation and enhance patient safety, and it was observed that the reciprocal motion of the capsule can help make the intestine stretch open to reduce the friction in narrow tubular environments compared with the DMA and CRMA methods, in this work, we apply the RRMA method in our autonomous magnetic navigation framework for safe and efficient actuation of the capsule.

\subsection{SMAL System and Algorithms}

\begin{figure*}[t]
\centering
\includegraphics[scale=1.0,angle=0,width=0.98\textwidth]{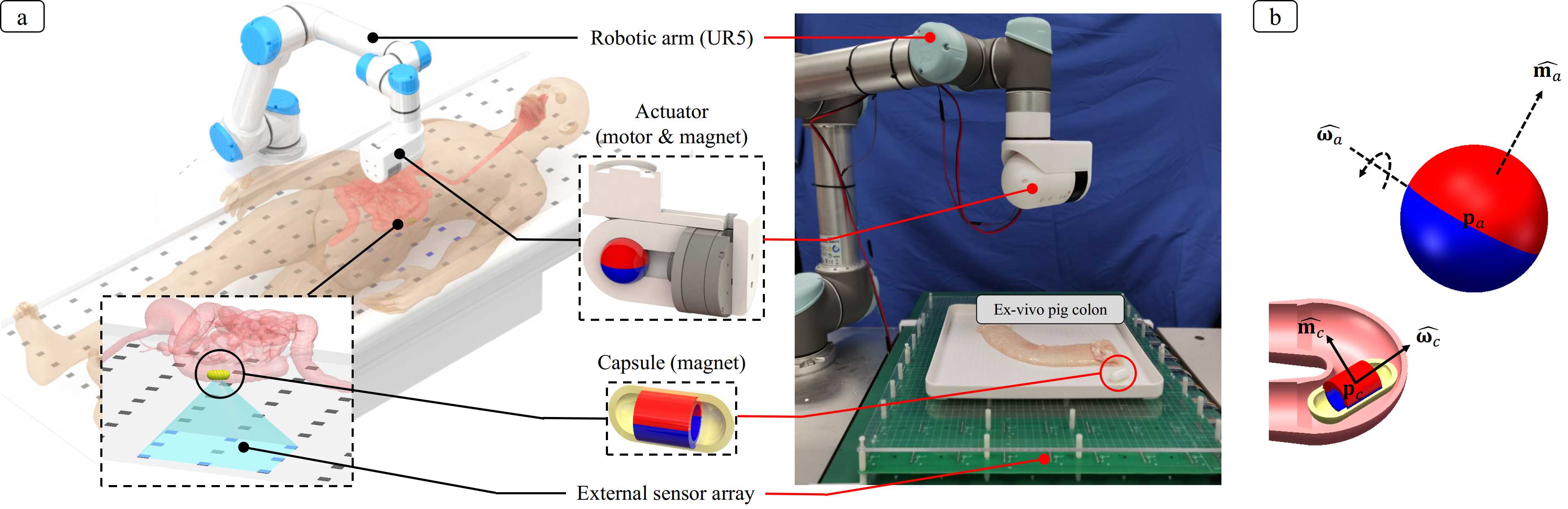}
\caption[The proposed autonomous navigation framework]{(a) Design and real-world system setup of our proposed robotic system for autonomous magnetic navigation of active WCE. The patient is required to swallow a capsule with an embedded magnetic ring and lie on an examination bed covered with a magnetic sensor array. The actuator magnet is controlled by a robotic arm and reciprocally rotates above the capsule to actuate it in the intestine. The capsule is localized based on the magnetic sensor feedback in real time. (b) The actuator with a unit magnetic moment of $\widehat{\pmb{m}}_{a}$ is rotating around $\widehat{\pmb{\omega}}_{a}$ at position $\mathbf{p}_{a}$ to actuate a capsule with a unit magnetic moment of $\widehat{\pmb{m}}_{c}$ at position $\mathbf{p}_{c}$ to rotate around $\widehat{\pmb{\omega}}_{c}$.}
\label{Fig_overview_systemconcepts}
\end{figure*}

The proposed autonomous navigation framework is implemented on a robotic system developed based on our previous work in \cite{xu2020novelsystem,xu2020improved,xu2021reciprocally} to allow closed-loop SMAL of a capsule. The design of the system is illustrated in Fig. \ref{Fig_overview_systemconcepts}(a), which uses an external spherical magnetic actuator controlled by a robotic arm to actuate a magnetic capsule inside the intestine, and the capsule is tracked by an external sensor array placed on the examination bed. This SMAL system only relies on the magnetic sensor data, which is not limited by the line-of-sight compared with visual sensor-based systems \cite{mahoney2016five}. Also, the use of the external sensor array for capsule localization can save the internal space of the capsule and reduce the power consumption compared with internal sensor-based systems \cite{pittiglio2019magnetic}\cite{barducci2019adaptive}.
As illustrated in Fig. \ref{Fig_overview_systemconcepts}(b), the center line of the magnetic ring embedded in the capsule coincides with the principal axis of the capsule. The 6-D poses of the capsule and the actuator can be represented by their positions $\mathbf{p}_{c}$, $\mathbf{p}_{a}$, unit magnetic moments $\widehat{\pmb{m}}_{c}$, $\widehat{\pmb{m}}_{a}$, and unit rotation axes (heading directions) $\widehat{\pmb{\omega}}_{c}$, $\widehat{\pmb{\omega}}_{a}$ \cite{xu2020improved}.

\subsubsection{Magnetic localization algorithm}

In order to track the capsule in a large workspace, we adopt the adaptive magnetic localization algorithm in \cite{xu2021adaptive} to estimate the 6-D pose of the capsule in real time. As outlined in Algorithm \ref{Alg_Localization}, the capsule's 5-D pose is first initialized based on the measurements of all the sensors using the multiple objects tracking (MOT) method \cite{song2016multiple} (line 1). Subsequently, a sensor sub-array with the optimal layout is adaptively selected and activated from the entire sensor array based on the capsule's position to improve the localization accuracy and update frequency (line 3-4). Then, after the 5-D pose of the capsule  ($\mathbf{p}_{c},\widehat{\pmb{m}}_{c}$) is estimated using the MOT method (line 5), the heading direction of the capsule $\widehat{\pmb{\omega}}_{c}$ is estimated using the normal vector fitting (NVF) method \cite{xu2020improved}.

\begin{algorithm}[t] 
\caption{Magnetic Localization Algorithm}
\label{Alg_Localization}
\KwIn{{magnetic field measurements from the external sensor array $\mathbf{B}^{(t)}$, $t=1,2,\cdots$}}
\KwOut{capsule's pose $(\mathbf{p}_{c}^{(t)},\widehat{\pmb{m}}_{c}^{(t)},{\widehat{\pmb{\omega}}_{c}^{(t)}})$ and velocity $\dot{\mathbf{p}}_{c}^{(t)}$, $t=1,2,\cdots$}
Initialize the 5-D pose of the capsule $(\mathbf{p}_{c}^{(0)}$, ${\widehat{\pmb{m}}_{c}^{(0)}})$ from $\mathbf{B}$ by the multiple objects tracking (MOT) algorithm\;
\For{$t=1, 2, \cdots$}
{
Select a sensor sub-array based on $\mathbf{p}_{c}^{(t-1)}$\;
Obtain magnetic field measurements from the selected sensors $\mathbf{B}_1^{(t)}  \subset \mathbf{B}^{(t)}$\;
Solve the current 5-D pose of the capsule $(\mathbf{p}_{c}^{(t)}$, ${\widehat{\pmb{m}}_{c}^{(t)}})$ from $\mathbf{B}_1^{(t)}$ by the multiple objects tracking (MOT) algorithm\;
Solve the capsule's heading direction $\widehat{\pmb{\omega}}_{c}^{(t)}$ from the 5-D pose sequence by the normal vector fitting (NVF) method\;
Estimate the capsule's velocity $\dot{\mathbf{p}}_{c}^{(t)}$\;
}
\Return {capsule's pose and velocity} $\mathbf{p}_{c}^{(t)},\widehat{\pmb{m}}_{c}^{(t)},{\widehat{\pmb{\omega}}_{c}^{(t)}},\dot{\mathbf{p}}_{c}^{(t)}$, $t=1,2,\cdots$\;
\end{algorithm}

\subsubsection{Automatic propulsion algorithm}

In order to control the movement of the actuator for the AP task, we use Algorithm \ref{Alg_AP} to calculate the desired pose of the actuator given the estimated capsule pose ($\mathbf{p}_{c}$, $\widehat{\pmb{m}}_{c}$, ${\widehat{\pmb{\omega}}_{c}}$) and velocity $\dot{\mathbf{p}}_{c}$. Based on the method presented in our previous work \cite{xu2021adaptive}, we adaptively change the actuator's heading direction according to the estimated moving speed of the capsule $v_{c}$ to efficiently and robustly propel the capsule in complex-shaped environments (line 2-3). However, different from \cite{xu2021adaptive} that uses CRMA, we employ RRMA \cite{xu2021reciprocally} in this work to improve patient safety and reduce environmental resistance in the narrow tubular environment (line 5).

\begin{algorithm}[t] 
\caption{Automatic Propulsion Algorithm}
\label{Alg_AP}
\KwIn{{capsule's pose ($\mathbf{p}_{c}^{(t)}$, $\widehat{\pmb{m}}_{c}^{(t)}$, $\widehat{\pmb{\omega}}_{c}^{(t)}$) and velocity $\dot{\mathbf{p}}_{c}^{(t)}$, $t=1,2,\cdots$}}
\KwOut{actuator's pose ($\mathbf{p}_{a}^{(t)}$, $\widehat{\pmb{m}}_{a}^{(t)}$, $\widehat{\pmb{\omega}}_{a}^{(t)}$), $t=1,2,\cdots$}
\For{$t=1, 2, \cdots$}
{Calculate the capsule's moving speed $v_{c}^{(t)}$\;
Calculate the desired heading direction of the  actuator $\widehat{\pmb{\omega}}_{a}^{(t)}$ based on $v_{c}^{(t)}$\;
Calculate the desired actuator position $\mathbf{p}_{a}^{(t)}$ based on the rotating magnetic actuation model\;
Determine $\widehat{\pmb{m}}_{a}^{(t)}$ by rotating the actuator with the RRMA method\;
}
\Return {actuator's pose} ($\mathbf{p}_{a}^{(t)}$, $\widehat{\pmb{m}}_{a}^{(t)}$, $\widehat{\pmb{\omega}}_{a}^{(t)}$), $t=1,2,\cdots$\;
\end{algorithm}

\subsubsection{Generation of the trajectory of the environment}
After the AP step is finished, the Gaussian Mixture Model (GMM) based Expectation Maximization (EM) algorithm is used to cluster the points in the trajectory \cite{xu2020improved}, and a smooth trajectory through these points is generated by cubic spline interpolation to represent the tubular environment as $\mathbf{p}_{traj}(s)$, $s \in [0.0,1.0]$, where $\mathbf{p}_{traj}(0)$ and $\mathbf{p}_{traj}(1)$ represent the first and last points on the trajectory, respectively. The user is allowed to select any point on the trajectory to carry out repeated inspections in the following step.

\subsubsection{Trajectory following algorithm}
Finally, during the TF step, given the capsule's pose and velocity ($\mathbf{p}_{c}$, ${\widehat{\pmb{\omega}}_{c}}$, $\dot{\mathbf{p}}_{c}$), intestinal trajectory $\mathbf{p}_{traj}$, and user-selected goal point $\mathbf{g}$, the system automatically actuates the capsule to reach the goal using the algorithm outlined in Algorithm \ref{Alg_TF}. First, the desired trajectory is obtained by truncating the entire intestinal trajectory between the goal point  $\mathbf{g}$ and the closest point to the current capsule position $\mathbf{p}_{c}$ (line 2). Then, the desired pose of the actuator is calculated using the TF algorithm in \cite{xu2021reciprocallyTF} based on the robust multi-stage model predictive controller (RMMPC) under RRMA (line 3-8).

\begin{algorithm}[t] 
\label{Alg_TF}
\caption{Trajectory Following Algorithm}
\KwIn{capsule's pose ($\mathbf{p}_{c}^{(t)}$, $\widehat{\pmb{m}}_{c}^{(t)}$, $\widehat{\pmb{\omega}}_{c}^{(t)}$) and velocity $\dot{\mathbf{p}}_{c}^{(t)}$, $t=1,2,\cdots$, pre-defined trajectory $\mathbf{p}_{traj}$, and user-selected goal point $\mathbf{g}$}
\KwOut{actuator's pose ($\mathbf{p}_{a}^{(t)}$, $\widehat{\pmb{m}}_{a}^{(t)}$, $\widehat{\pmb{\omega}}_{a}^{(t)}$), $t=1,2,\cdots$}
\For{$t=1, 2, \cdots$}
{Obtain the desired trajectory $\mathbf{p}'_{traj}$ by truncating $\mathbf{p}_{traj}$ between $\mathbf{g}$ and the closest point to $\mathbf{p}_{c}^{(t)}$\;
Calculate the desired position and velocity sequences of the capsule $\mathbf{p}_{d,0}\cdots\mathbf{p}_{d,N}$, $\dot{\mathbf{p}}_{d,0}\cdots\dot{\mathbf{p}}_{d,N}$ within the prediction horizon $N$ based on $\mathbf{p}_{c}^{(t)}$, $\dot{\mathbf{p}}_{c}^{(t)}$ and $\mathbf{p}'_{traj}$\;
Calculate the desired magnetic force sequence $\mathbf{f}_{d,0}\cdots\mathbf{f}_{d,N-1}$ based on the robust multi-stage model predictive controller (RMMPC)\;
Obtain the desired heading direction $\widehat{\pmb{\omega}_{dc}}=\frac{\dot{\mathbf{p}}_{d,0}}{\|\dot{\mathbf{p}}_{d,0}\|}$\;
Obtain the desired magnetic force $\mathbf{f}_{d}=\mathbf{f}_{d,0}$\;
Calculate the next heading direction $\widehat{\pmb{\omega}_{nc}}$ based on ${\widehat{\pmb{\omega}}_{dc}}$ and ${\widehat{\pmb{\omega}}_{c}^{(t)}}$\;
Calculate the desired pose of the actuator ($\mathbf{p}_{a}^{(t)}$, $\widehat{\pmb{m}}_{a}^{(t)}$, $\widehat{\pmb{\omega}}_{a}^{(t)}$) based on the RRMA method\;
}
\Return{actuator's pose} ($\mathbf{p}_{a}^{(t)}$, $\widehat{\pmb{m}}_{a}^{(t)}$, $\widehat{\pmb{\omega}}_{a}^{(t)}$), $t=1,2,\cdots$\;
\end{algorithm}

\section{EXPERIMENTS AND RESULTS}
\label{L_experiment}
In order to investigate the feasibility of the proposed autonomous navigation framework, we conducted a set of real-world experiments to study the navigation performance of our system in unknown, complex tubular environments. The results are reported and discussed in this section.
\subsection{System Setup}
\label{L_experiment_setup}

The real-world system setup for our navigation experiments can be seen in Fig. \ref{Fig_overview_systemconcepts}. An actuator consisting of a motor (RMDL-90, GYEMS) and a spherical permanent magnet (diameter $50mm$, NdFeB, N42 grade) is installed at the end-effector of a 6-DoF serial robotic manipulator (5-kg payload, UR5, Universal Robots). The capsule (diameter $16mm$, length $35mm$) is comprised of a 3D-printed shell (Polylactic Acid, UP300 3D printer, Tiertime) and a permanent magnetic ring (outer diameter $12.8mm$, inner diameter $9mm$, and length $15mm$, NdFeB, N38SH grade). The large external sensor array comprises $80$ three-axis magnetic sensors (MPU9250, InvenSense) arranged in an $8 \times 10$ grid with a spacing of $6cm$ to cover the entire abdominal region of the patient. The output frequency of each sensor is $100Hz$. $10$ USB-I2C adaptors (Ginkgo USB-I2C, Viewtool), a USB-CAN adaptor (Ginkgo USB-CAN, Viewtool) and a network cable are used for data transmission. All proposed algorithms are implemented with Python and run on a desktop (Intel i7-7820X, 32GB RAM, Win10).

\subsection{Evaluation of Different Magnetic Actuation Methods in the AP Task}
\label{L_experiment_AP}

\begin{figure}[t]
\setlength{\abovecaptionskip}{-0.5cm}
\centering
\includegraphics[scale=1.0,angle=0,width=0.49\textwidth]{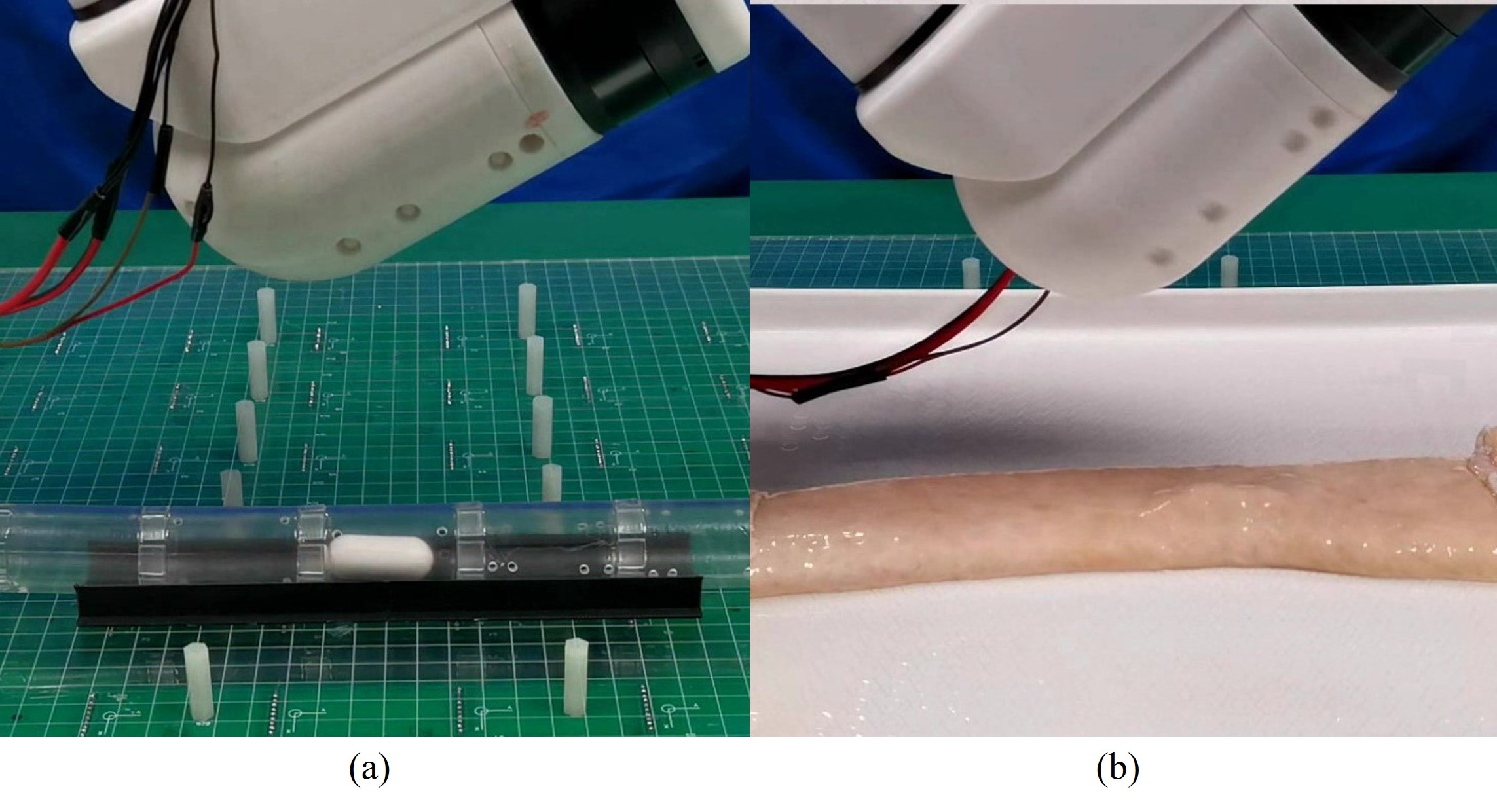}
\caption[Two environments for evaluating the performance of different magnetic actuation methods in the automatic propulsion and trajectory following tasks]{The experiments conducted in (a) a PVC tube and (b) an ex-vivo pig colon to evaluate the performance of different magnetic actuation methods in the AP and TF tasks.}
\label{Fig_experiment_3MA}
\end{figure}

As the first step in our autonomous navigation framework, it is important to ensure safe and efficient AP of the capsule to explore the unknown tubular environment. We first compare the performance of the system in the AP task when using three different magnetic actuation methods, i.e., DMA, CRMA and RRMA. The experiments were conducted in a PVC tube and an ex-vivo pig colon, as shown in Fig. \ref{Fig_experiment_3MA}. We carried out five instances for each test, and the position of the actuator relative to the capsule is kept unchanged during all the experiments.
The results are presented in Table \ref{T_AP} and Movie S1 (please refer to section \ref{L_experiment_demo}). We found that all the three magnetic actuation methods can successfully propel the capsule through the straight PVC tube (see Fig. \ref{Fig_experiment_3MA} (a)) in the five trials; however, the moving speed of the capsule using rotating magnetic actuation methods is much faster ($\sim 10$ times) than that under dragging-based actuation. This is mainly because the rotating magnetic actuation can reduce the friction between the capsule and the tube to accelerate the exploration in the unknown tubular environment. In the experiments in the ex-vivo pig colon (see Fig. \ref{Fig_experiment_3MA}(b)), since the friction becomes larger, the capsule cannot be effectively propelled using DMA. While CRMA has a lower success rate of $40\%$, RRMA can still reach a success rate of $100\%$ to advance the capsule through the environment with a speed of about $2.5 mm/s$. It is observed in the experiments that the CRMA can occasionally cause the malrotation of the intestinal wall, which may increase the environmental resistance and hinder the advancement of the capsule. Instead, RRMA does not cause the malrotation of the intestine in all the experiments, which shows the potential of RRMA to reduce patient discomfort and improve patient safety \cite{xu2021reciprocally}. Moreover, since the reciprocal rotation of the capsule helps make the intestine stretch open during propulsion, the environmental resistance can be reduced to make the capsule easily pass through the narrow tubular environment, which is important to improve the overall efficiency of the autonomous navigation.

\begin{table}[t] \renewcommand\arraystretch{1.2} \small
\centering
\caption{Evaluation of Different Magnetic Actuation Methods in The AP Task}
\begin{tabular}{|p{2.2cm}<{\centering}|p{1.6cm}<{\centering}|p{1.4cm}<{\centering}|p{1.8cm}<{\centering}|}
\hline
\multirow{2}{*}{\textbf{Environment}} & \multirow{2}{*}{\tabincell{c}{\textbf{Actuation} \\ \textbf{method}}} & \multirow{2}{*}{\textbf{\tabincell{c}{Success\\ rate}}} & \multirow{2}{*}{\tabincell{c}{\textbf{Average speed} \\ $\mathbf{(mm/s)}$}} \\
{} & {} & {} & {} \\
\hline
\multirow{3}{*}{\tabincell{c}{In a PVC tube \\ (length: $180 mm$)}} & DMA & {$100 \%$} & {$0.81$} \\
\cline{2-4}
{} & {CRMA} & {$100 \%$} & {$8.41$} \\
\cline{2-4}
{} & {RRMA} & {$\mathbf{100 \%}$} & {$\mathbf{8.70}$} \\
\hline
\multirow{3}{*}{\tabincell{c}{In an ex-vivo pig \\ colon (length:\\$155 mm$)}} & DMA & {$0 \%$} & {$-$} \\
\cline{2-4}
{} & {CRMA} & {$40 \%$} & {$2.58$} \\
\cline{2-4}
{} & {RRMA} & {$\mathbf{100 \%}$} & {$\mathbf{2.48}$} \\
\hline
\end{tabular}
\label{T_AP}
\end{table}

\subsection{Evaluation of Different Magnetic Actuation Methods in the TF Task}
\label{L_experiment_TF}

We further evaluate the performance of the system using different magnetic actuation methods in the TF task in the same environments as shown in Fig. \ref{Fig_experiment_3MA}. Five instances of each test were conducted, with the relative pose of the actuator to the capsule set the same as in the AP task. The desired trajectory for each tubular environment is manually specified.
As can be seen in Table \ref{T_TF} and Movie S1 (please refer to section \ref{L_experiment_demo}), the DMA method has the worst tracking accuracy in the PVC tube and cannot finish the trajectory following task in the ex-vivo pig colon. In contrast, we found that the CRMA and RRMA methods can successfully actuate the capsule to follow the pre-defined trajectories in both environments, and RRMA achieves the best tracking accuracy. This shows that the DMA method is difficult to overcome the large friction in narrow tubular environments to complete the TF task, while the rotating-based actuation methods can inherently reduce the friction. In addition, the continuous rotation would cause a shift in the capsule position during the actuation, which would reduce the tracking accuracy in the TF task. The results suggest that the RRMA method can realize both robust and accurate trajectory following of the capsule given the desired trajectory, which has the potential to realize repeated inspection of suspicious points in the intestine.

According to the above analysis, it can be concluded that the RRMA method can achieve the best performance in both the AP and TF tasks in terms of propulsion efficiency and tracking accuracy in the intestinal environment and has the potential to improve patient safety during the procedure.

\begin{table}[t] \renewcommand\arraystretch{1.2} \small
\centering
\caption{Evaluation of Different Magnetic Actuation Methods in The TF Task}
\begin{tabular}{|p{2.2cm}<{\centering}|p{1.7cm}<{\centering}|p{1.4cm}<{\centering}|p{1.7cm}<{\centering}|}
\hline
\multirow{2}{*}{\textbf{Environment}} & \multirow{2}{*}{\tabincell{c}{\textbf{Actuation} \\ \textbf{method}}} & \multirow{2}{*}{\textbf{\tabincell{c}{Success\\ rate}}} & \multirow{2}{*}{\tabincell{c}{\textbf{Accuracy} \\ $\mathbf{(mm)}$}} \\
{} & {} & {} & {} \\
\hline
\multirow{3}{*}{\tabincell{c}{In a {PVC} tube \\ (length: $180 mm$)}} & {DMA} & {$100 \%$} & {$15.3 \pm 2.8$} \\
\cline{2-4}
{} & {CRMA} & {$100 \%$} & {$9.8 \pm 2.2$} \\
\cline{2-4}
{} & {RRMA} & {$\mathbf{100 \%}$} & {$\mathbf{3.8 \pm 2.5}$} \\
\hline
\multirow{3}{*}{\tabincell{c}{In an ex-vivo pig \\ colon (length:\\$155 mm$)}} & {DMA} & {$0 \%$} & {$-$} \\
\cline{2-4}
{} & {CRMA} & {$100 \%$} & {$9.2 \pm 3.8$} \\
\cline{2-4}
{} & {RRMA} & {$\mathbf{100 \%}$} & {$\mathbf{7.9 \pm 4.2}$} \\
\hline
\end{tabular}
\label{T_TF}
\end{table}

\subsection{Evaluation of the Overall Autonomous Magnetic Navigation Framework}
\label{L_experiment_navigation}

\begin{figure*}[t]
\setlength{\abovecaptionskip}{-0.0cm}
\centering
\includegraphics[scale=1.0,angle=0,width=0.98\textwidth]{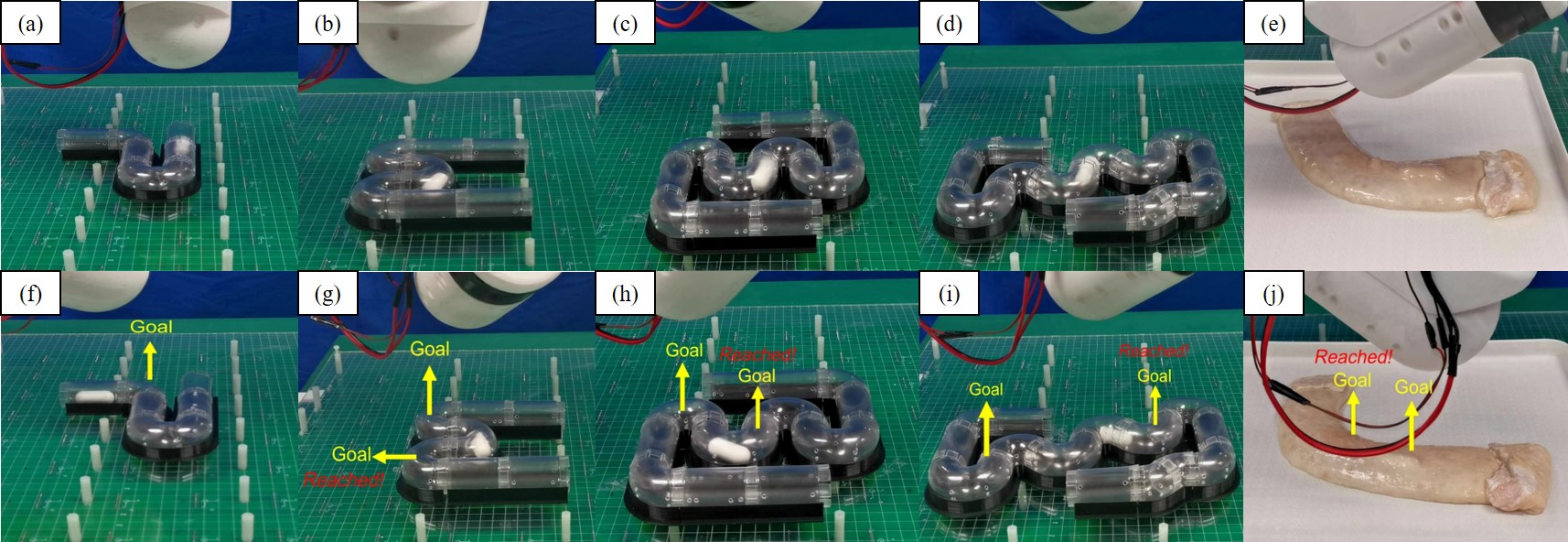}
\caption[Demos of the overall autonomous magnetic navigation framework]{Navigation experiments conducted in five different tubular environments to evaluate the performance of the overall autonomous magnetic navigation framework. (a-e) show the experiments during the ``insertion" step, when the capsule is propelled through four PVC tubes with different shapes and lengths and an ex-vivo pig colon, respectively. (f-j) show the experiments during the ``withdrawal" step, when the capsule is navigated towards the given goal points (marked in yellow) in the four PVC tubes and an ex-vivo pig colon, respectively.}
\label{Fig_experiment_5APTF}
\end{figure*}

In order to evaluate the performance of the overall autonomous magnetic navigation framework, we conducted the navigation experiments in five tubular environments with different complexities, including four PVC tubes with different shapes and lengths and an ex-vivo pig colon. Each experiment is composed of two steps, the ``insertion" step and ``withdrawal" step, as described in Section~II-A. During the ``insertion" step, AP under RRMA is applied to autonomously advance the capsule through the unknown tubular environment, and a smooth trajectory is generated. Then, several points are manually selected on the trajectory as suspicious lesions that need to be revisited by the capsule. Three different approaches are implemented and compared for the ``withdrawal" step, including \textit{tele-operation}, \textit{backward AP} and \textit{TF}. In the tele-operation mode, the user can observe the desired trajectory generated by AP and manually send motion commands to navigate the capsule. In the backward AP mode, the system will automatically propel the capsule along the inverse direction of the trajectory and does not use the trajectory generated by the forward AP. The TF mode is to automatically actuate the capsule using the method described in Section~II-C. The experiment results are summarized in Table \ref{T_navigation}.

We first evaluate the system performance in the ``insertion" step using the proposed AP method in different environments. As shown in Fig. \ref{Fig_experiment_5APTF}(a-e) and Table \ref{T_navigation}, we found that the capsule can be efficiently propelled through all the five unknown tubular environments with a moving speed ranging from $1mm/s$ to $5mm/s$, and the system can successfully generate a smooth trajectory to represent each tubular environment.

Then, we take a look at the ``withdrawal" step (see Fig. \ref{Fig_experiment_5APTF}(f-j)) to assess the system's ability to allow accurate navigation towards suspicious lesions. One goal point is set in Tube No.1 as it has a relatively short length. In the other tubular environments, two goal points are set and the capsule is required to reach the first goal, then reach the second goal, and finally revisit the first goal. We evaluate the three methods (i.e., tele-operation, backward AP and TF) based on the accuracy (distance between the final position of the capsule and the goal), repeatability (distance between the two final positions of the capsule navigating towards the same goal twice) and the average speed to achieve the goal. The quantitative results are summarized in Table \ref{T_navigation}. We found that the three methods show similar accuracy in all five environments, with a tracking error of $2 \sim 3 mm$. However, the repeatability of the manual operation is the worst among all three methods, especially in the ex-vivo pig colon, which may be due to the low accuracy of visual localization. Also, the tele-operation is tedious and time-consuming as it requires the user to continuously observe the screen and click the control panel. The backward AP method can automatically propel the capsule backward, but it can only make decisions based on the current environmental information and thus cannot quickly navigate towards the goal. In contrast, the TF method achieves the best repeatability and efficiency in all five tubular environments, as it can take advantage of the knowledge of the intestine gained during the ``insertion" step to better control the movement of the capsule. The slight deterioration in the tracking accuracy of TF in some experiments (e.g., in tubes No.2 and No.4) can be attributed to the fact that the TF algorithm will produce an overshoot to reach the target as soon as possible in the navigation process, but this slight decrease in accuracy (less than $1 mm$) does not affect the overall navigation performance.

\begin{table*}[tb] \renewcommand\arraystretch{1.2} \small
\centering
\caption{Evaluation of the Overall Autonomous Navigation Framework in Different Tubular Environments}
\begin{tabular}{|p{2.6cm}<{\centering}|p{1.8cm}<{\centering}|p{2.0cm}<{\centering}|p{2.2cm}<{\centering}|p{2.0cm}<{\centering}|p{2.0cm}<{\centering}|p{1.8cm}<{\centering}|}
\hline
\multirow{3}{*}{\textbf{Environment}} & \multicolumn{2}{c|}{\textbf{``Insertion" step}} & \multicolumn{4}{c|}{\textbf{``Withdrawal" step}} \\
\cline{2-7}
{} & \multirow{2}{*}{\textbf{Method}} & \multirow{2}{*}{\tabincell{c}{\textbf{Average speed} \\ $\mathbf{(mm/s)}$}} & \multirow{2}{*}{\textbf{Method}} & \multirow{2}{*}{\tabincell{c}{\textbf{Accuracy} \\ $\mathbf{(mm)}$}} & \multirow{2}{*}{\tabincell{c}{\textbf{Repeatability} \\ $\mathbf{(mm)}$}} & \multirow{2}{*}{\tabincell{c}{\textbf{Average speed} \\ $\mathbf{(mm/s)}$}} \\
{} & {} & {} & {} & {} & {} & {} \\
\hline
\multirow{3}{*}{\tabincell{c}{Tube No.1\\(length: $224mm$)}} & \multirow{15}{*}{AP} & \multirow{3}{*}{$2.29$} & {Tele-operation} & {$2.9 \pm 2.1$} & \multirow{3}{*}{$-$} & {$0.66$} \\
\cline{4-5}\cline{7-7}
{} & {} & {} & {Backward AP} & {$2.7 \pm 1.9$} & {} & {$0.95$} \\
\cline{4-5}\cline{7-7}
{} & {} & {} & {TF} & {$\mathbf{2.8 \pm 2.3}$} & {} & {$\mathbf{2.74}$} \\
\cline{1-1}\cline{3-7}
\multirow{3}{*}{\tabincell{c}{Tube No.2\\(length: $378mm$)}} & {} & \multirow{3}{*}{$2.80$} & {Tele-operation} & {$3.1 \pm 2.5$} & {$6.8$} & {$0.50$} \\
\cline{4-7}
{} & {} & {} & {Backward AP} & {$3.3 \pm 2.8$} & {$7.8$} & {$0.88$} \\
\cline{4-7}
{} & {} & {} & {TF} & {$\mathbf{3.7 \pm 3.1}$} & {$\mathbf{5.3}$} & {$\mathbf{1.68}$} \\
\cline{1-1}\cline{3-7}
\multirow{3}{*}{\tabincell{c}{Tube No.3\\(length: $564 mm$)}} & {} & \multirow{3}{*}{$1.83$} & {Tele-operation} & {$2.1 \pm 1.6$} & {$12.4$} & {$0.74$} \\
\cline{4-7}
{} & {} & {} & {Backward AP} & {$2.4 \pm 1.8$} & {$11.28$} & {$0.77$} \\
\cline{4-7}
{} & {} & {} & {TF} & {$\mathbf{2.2 \pm 1.6}$} & {$\mathbf{5.6}$} & {$\mathbf{1.60}$} \\
\cline{1-1}\cline{3-7}
\multirow{3}{*}{\tabincell{c}{Tube No.4\\(length: $678mm$)}} & {} & \multirow{3}{*}{$2.30$} & {Tele-operation} & {$2.4 \pm 1.8$} & {$12.6$} & {$0.71$} \\
\cline{4-7}
{} & {} & {} & {Backward AP} & {$2.5 \pm 1.7$} & {$12.2$} & {$0.78$} \\
\cline{4-7}
{} & {} & {} & {TF} & {$\mathbf{3.2 \pm 2.7}$} & {$\mathbf{9.5}$} & {$\mathbf{1.88}$} \\
\cline{1-1}\cline{3-7}
\multirow{3}{*}{\tabincell{c}{Ex-vivo pig colon \\(length: $200$ $mm$)}} & {} & \multirow{3}{*}{$5.10$} & {Tele-operation} & {$4.7 \pm 3.1$} & {$5.6$} & {$0.52$} \\
\cline{4-7}
{} & {} & {} & {Backward AP} & {$3.4 \pm 3.1$} & {$2.7$} & {$0.76$} \\
\cline{4-7}
{} & {} & {} & {TF} & {$\mathbf{3.6 \pm 3.1}$} & {$\mathbf{1.5}$} & {$\mathbf{0.78}$} \\
\hline
\end{tabular}
\label{T_navigation}
\end{table*}

\subsection{Video Demonstration}
\label{L_experiment_demo}

Video demonstration of the three magnetic actuation methods in the AP and TF tasks can be seen in Movie S1\footnote{Available at \url{https://youtu.be/cnBGfmFjnTk} or \url{https://www.bilibili.com/video/BV1Q64y1B7G3/}}. Video demonstration of the autonomous magnetic navigation in different tubular environments can be seen in Movie S2\footnote{Available at \url{https://youtu.be/X9ESK8J4KUI} or \url{https://www.bilibili.com/video/BV12f4y1G7yn/}}.

\section{CONCLUSIONS}
\label{L_conclusion}

In this paper, we present a framework for autonomous magnetic navigation of active WCE in an unknown tubular environment inspired by the procedures of conventional colonoscopy, and describe its potential use in a clinical setting. The automatic propulsion (AP) and trajectory following (TF) of a magnetic capsule are performed by a robotic system to mimic the ``insertion" and ``withdrawal" techniques performed by an expert physician in routine colonoscopy, in order to allow for efficient and accurate navigation in unknown intestinal environments. Our method is implemented on a real robotic system and validated in extensive navigation experiments in phantoms and an ex-vivo pig colon. Our results preliminarily demonstrate that the reciprocally rotating magnetic actuation method used in our system can achieve satisfactory performance in both the AP and TF tasks, and the overall framework for autonomous magnetic navigation of active WCE can effectively navigate the capsule towards desired positions in unknown, complex tubular environments with minimal user effort, which has the potential to reduce the examination time and improve the diagnostic outcome for WCE.

This technology can potentially advance the field of medical robotics by providing a general solution to autonomous positioning of a medical robot in an unknown tubular environment, which may improve the usability and clinical adaptation of active locomotion technologies for different medical applications.

\bibliographystyle{IEEEtran}
\bibliography{root}

\end{document}